\documentclass{article} % For LaTeX2e
\pdfoutput=1
\usepackage{iclr2016_conference,times}
\usepackage{url}
\usepackage[utf8]{inputenc}

\usepackage{graphicx}
\usepackage{caption}
\usepackage{subcaption}
\usepackage{stmaryrd}
\usepackage{amsfonts}
\usepackage{amsmath}
\usepackage{wrapfig}
\usepackage{sidecap}
\sidecaptionvpos{figure}{t}

 \usepackage{algorithm}
 \usepackage{algorithmic}

\long\def\comment#1{}

\title{Geodesics of learned representations}

\author{Olivier J.~Hénaff \& Eero P.~Simoncelli \\
Howard Hughes Medical Institute, \\
Center for Neural Science and \\
Courant Institute of Mathematical Sciences \\
New York University\\
New York, NY 10003, USA \\
\texttt{\{henaff, eero\}@cns.nyu.edu}}

\iclrfinalcopy % Uncomment for camera-ready version

\begin{document}

\maketitle

\begin{abstract}
We develop a new method for visualizing and refining the invariances of learned representations. Specifically, we test for a general form of invariance, linearization, in which the action of a transformation is confined to a low-dimensional subspace. Given two reference images (typically, differing by some transformation), we synthesize a sequence of images lying on a path between them that is of minimal length in the space of the representation (a ``representational geodesic"). If the transformation relating the two reference images is linearized by the representation, this sequence should follow the gradual evolution of this transformation. We use this method to assess the invariance properties of a state-of-the-art image classification network and find that geodesics generated for image pairs differing by translation, rotation, and dilation do not evolve according to their associated transformations. Our method also suggests a remedy for these failures, and following this prescription, we show that the modified representation is able to linearize a variety of geometric image transformations.
\end{abstract}

\section{Introduction}

A fundamental requirement of pattern recognition is the ability to ignore irrelevant variations in the input \citep{Duda:2001}. Most visual recognition problems are thwarted by variations in position, size, pose, lighting, and other viewing conditions that can bring objects from different classes closer, while increasing within-class variability \citep{DiCarlo:2007hs}, and the construction of representations that are invariant to these variations remains an active area of research. Recent examples of learned visual representations have proven highly effective for recognition \citep{Krizhevsky:2012wl}, but a precise understanding of exactly what they represent remains elusive. And although these representations are hypothesized to be invariant to various identity-preserving deformations, apart from a few exceptions, these claims are rarely tested directly \citep{Kavukcuoglu:2009ux}.

Image synthesis provides a powerful methodology for examining the invariances of arbitrary representations. It has been used to explore and refine texture models, incrementally augmenting the representation with new statistical constraints until images synthesized with matching parameters are indistinguishable to human observers \citep{Portilla:2000gl}. When applied to deep recognition networks, synthesis has revealed failures in the form of ``adversarial examples": images that appear entirely different to a human observer, and yet are identified by the network as belonging to the same category \citep{Szegedy:2013vw}. In these cases, samples from the equivalence class of images that map to the same representation vector provide a means of verifying or falsifying the hypothesis that the invariances of the representation are also invariances for human observers.
 
\comment{By sampling from the equivalence class of images that map to the same representation, synthesis exposes the invariances of a model through the diversity of synthetic images. That said, it has mainly be used successfully for exploring \textit{totally-ordered} pairs of models, in which one model discards more information than the other. \citet{Portilla:2000gl} used synthesis for exploring a set of nested texture models, incrementally augmenting the representation until it produced perceptually indistinguishable synthetic images, or visual metamers. Similarly, \citet{Freeman:2011gl} used synthesis in a set of models parametrized by receptive field size, effectively controlling the amount of discarded information. In their experiment, they progressively shrunk the amount of discarded information until observers could no longer distinguish between synthetic images produced by the model. In both cases, the total-order on the class of models being studied enabled them to find, out of the subset of models that synthesize visual metamers, the model which is most invariant.}

\begin{figure}[t]%[!hb]
    \centering
    \includegraphics[width=1\textwidth]{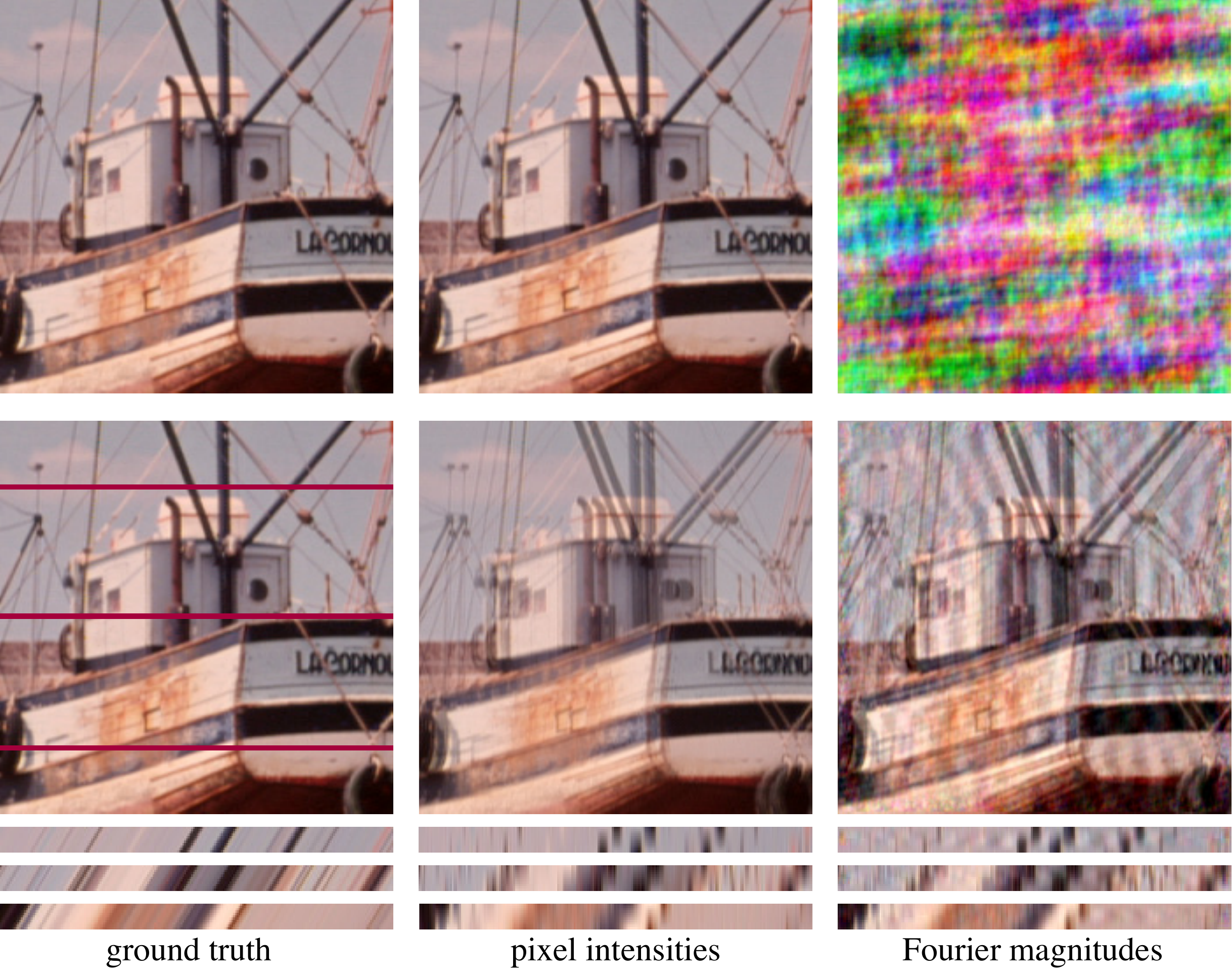}
    \caption{Geodesics can reveal either insufficient or excessive invariance, whereas synthesis reveals only the latter. {\bf Top}: images synthesized so that their representation is matched to that of the ground truth image (left). Middle image has matching pixel intensities (i.e., it is identical to the ground truth image) and right image has matching Fourier magnitudes. {\bf Bottom}: synthetic geodesic sequences connecting two translated copies of the same image, via different representations. Shown are the middle frame of each sequence, and below it, the temporal evolution of each row of pixels indicated by a horizontal red line. The ground truth transformation (left) is a translation (as can be seen from the diagonal lines in the temporal slice), but both geodesics deviate from the true transformation. The pixel representation fully constrains the image, and has no invariances, and thus the synthesized geodesic images are simply linearly interpolated between the initial and final images. The Fourier magnitudes, while translation-invariant, are also invariant to arbitrary phase perturbations, and the synthesized geodesic image contains Fourier components whose phases are shifted inconsistently.}
    \label{fig:toy_example}
\end{figure}

But the synthesis test, in which human observers try to discriminate synthesized images, is one-sided: failures (i.e.\ visually distinct images) can reveal {\em inappropriate} invariances of a representation, but successes can mask a lack of {\em desired} invariances. Consider the standard case of translation-invariance.  The Fourier amplitude spectrum (i.e., the set of magnitudes of Fourier transform coefficients) provides a well-known example of a translation-invariant representation, but it is invariant to far more than translations, and this is immediately revealed by a synthesis test (figure \ref{fig:toy_example}, top right).  On the other hand, simply representing an image with its raw pixel values (the identity representation) will trivially produce visually perfect synthetic examples (figure \ref{fig:toy_example}, top center) despite the fact that it has no invariance properties at all. 

We seek a more general method of evaluation that penalizes a model for discarding too much information (as with synthesis) but also for discarding too little information. Each of these failures can be seen as an inadequacy of the image {\em metric} induced by the representation. Specifically, an image representation deforms the input space, bringing some images closer to each other while spreading others out, and thus inducing a new metric in image space. We can expose properties of this image metric by generating a geodesic sequence of images. Specifically, given an initial and final image, we synthesize a sequence of images that follow a minimal-length path in the response space of the representation. In the absence of any other constraints, this path will be a straight line connecting the representations of the two images; more generally, it will be the \textit{straightest} path connecting the two points. In the case where the two images differ by a simple transformation (e.g.\ a translation, figure \ref{fig:toy_example}, left column) that is not linearized by the representation (i.e. mapped to the straightest path connecting the two representations), the geodesic will differ from the original transformation connecting the images (figure \ref{fig:toy_example}, middle column). Similarly, if the representation is invariant to many transformations, the geodesic may correspond to a path that uses a mixture of transformations, and thus differ from the ground truth path (figure \ref{fig:toy_example}, right column). As a result, by visualizing whether a representation has linearized the action of various deformations, representational geodesics can reveal both excessive and insufficient invariance in an image model. 
%It should be noted that linearization is a weak form of invariance, as it confines the effect of a transformation to linear subspace, which can then be easily projected out with a linear readout or classifier.
 
We develop an algorithm for synthesizing geodesic sequences for a representation, and use it to examine whether learned representations linearize various real-world transformations such as translation, rotation, and dilation. We find that a current state-of-the-art object recognition network fails to linearize these basic transformations. However, these failures point to a deficiency in the representation, leading to a simple way of improving it. We show that the improved representation is able to linearize a range of parametric transformations as well as generic distortions found in natural image sequences. 

%%%%%%%%%%%%%%%%%%%%%%%%%%%%%%%%%%%%%%%%%%%%%%%%%%%%%%%%%%%%%%%%%%%
\section{Synthesizing geodesic sequences}
\label{sec:methods}

Suppose we have an image representation, $y=f(x)$, where $x$ is the vector of image pixel intensities and $f(\cdot)$ a continuous function that maps it to an abstract vector-valued representation $y$ (e.g.\ the responses of an intermediate stage of a hierarchical neural network). Given initial and final images, we wish to synthesize a sequence of images that lies along the path of minimal length  in the representation space (a {\em representational geodesic}).  If the mapping is many-to-one (as is usually the case), this sequence of images is not unique.  We resolve this ambiguity by selecting the representational geodesic that is also of minimal length in the space of images (i.e.,  a {\em conditional geodesic} in image space).

\subsection{Objective function}

In order to generate such a sequence, we optimize an objective function that expresses a discrete approximation of the problem, directly in terms of images sampled along the path.
Given a desired sequence length $N$ and  initial and final images, $\{x_0, x_N\}$, we wish to synthesize a sequence of images, $ \gamma = \{x_n ; n=0\ldots N\}$, lying along a geodesic in representation space. The representational path length is  
\[ 
L[ f( \gamma ) ] = \sum_{n=1}^{N} \left\Vert f(x_n) - f(x_{n-1}) \right\Vert_2
\] 
which is bounded by the representational energy
\[ E[ f( \gamma ) ] = \sum_{n=1}^{N} \left\Vert f(x_n) - f(x_{n-1}) \right\Vert^2_2 \]
thanks to the Cauchy-Schwartz inequality
\[ L[ f( \gamma ) ]^2 \leq N E[ f( \gamma ) ] \]
with equality if and only if the representations are equispaced, which is encouraged by minimizing the representational energy. As a result, a path that meets this condition (e.g. the red curve in figure \ref{fig:deviation}) while minimizing the representational energy $E[ f( \gamma ) ]$ is a representational geodesic.

When the mapping to representation space is many-to-one, there are many possible solutions to this problem. 
% In the absence of constraints, this geodesic is a sequence of vectors lying along the line between the initial and final representations:
% \[
% y_n  =  \frac{N-n}{N} y_0 + \frac{n}{N} y_N \quad n \in \llbracket 0,1,\dots  N \rrbracket 
% \]
% where $y_0 = f(x_0)$ and $y_N = f(x_N)$.
% We then define an objective that penalizes the deviation of the sequence of representations from this target set of equispaced vectors along the representation line:
% \[ 
% \En_{\mani}(x) = \sum_{n=1}^{N-1} \left\Vert f(x_n) - y_n \right\Vert^2_2 
% \]
To uniquely constrain the solution, we define an analogous energy term that ensures that this path is also of minimal length in the image domain
\[ E[ \gamma ] = \sum_{n=1}^{N} \left\Vert x_n - x_{n-1} \right\Vert^2_2 \]
% , and that we sample it uniformly.  Specifically, we penalize the maximal distance between successive images in the sequence:
% \[ 
% \En_{\cont}(x) = \max_{n\in \llbracket 1,2,\ldots N \rrbracket} \left\Vert x_n - x_{(n-1)} \right\Vert^2_2 
% \]
Since we are looking for the shortest path in image space that is also a geodesic in representation space, we minimize $E[ \gamma ]$  conditioned on the path also minimizing $E[ f( \gamma ) ]$. 
% \[ \gamma^* = \arg \min_{\gamma \in \arg \min E[ f( \gamma ) ] } E[ \gamma ]  \]
% \begin{equation*}
% \begin{aligned}
% & \underset{\gamma}{\text{minimize}}
% & & E[ \gamma ] \\
% & \text{subject to}
% & & \gamma \in \arg \min E[ f( \gamma ) ]
% \end{aligned}
% \end{equation*}
% Since we are looking for the shortest path in image space that is a geodesic in representation space, we would ideally minimize $E[ \gamma ]$  conditioned on the path minimizing $E[ f( \gamma ) ]$. In practice, we optimize a weighted sum of the two objective functions:
% \[ F( \gamma ) = E[ f( \gamma ) ] + \lambda \ E[ \gamma ] \]
% where the  weight $\lambda$ is set as high as possible, conditioned on the resulting representation error $E_\mani$ being approximately equal to that obtained for $\lambda = 0$. 
Furthermore, during the optimization we constrain image pixel intensities to the [0, 1] range.  

\subsection{Optimization}
We optimize this objective in three steps. First, we initialize the path with the minimum of $E[ \gamma ]$, which is simply a sequence of images that are linearly interpolated between the initial and final images. Next we minimize the representational geodesic objective $E[ f( \gamma ) ]$. Finally, we minimize the image-domain geodesic objective, conditioned on staying in the set of representational geodesics.

Minimizing the representational geodesic objective in the second step requires optimizing an image for its representation via a non-linear function, and thus shares much of the non-convexity found in training deep neural networks. In particular, the curvature of the energy surface can vary widely over the course of the optimization. For this reason, we used the Adam optimization method \citep{Kingma:2014us}, which scales gradients by a running estimate of their variance, providing robustness to these changes in the energy landscape. We run Adam, using the default parameters, for $10^4$ iterations to ensure that we reach the minimum of the representational geodesic cost. 

To optimize the image-domain geodesic objective while constraining the solution to remain in the set of representational geodesics, we start by computing a descent direction for the image-domain geodesic objective. We then project out the component of this direction that lies along the gradient of the representational geodesic objective. We take a step in that direction, then project back onto the set of representational geodesics by re-minimizing the representational geodesic cost (again using Adam), and repeat until convergence. We summarize our method with the following algorithm. 
\begin{algorithm}[h]
\caption*{Conditional geodesic computation } 
\begin{algorithmic}  

\vspace{0.1cm}

    \REQUIRE $f$: continuous mapping 
\vspace{0.1cm}
    \REQUIRE $x_0, x_N$: initial and final images 

\vspace{0.1cm}

\REQUIRE $N$: number of steps along geodesic path ($N = 10$ in all our experiments)

\vspace{0.1cm}

\REQUIRE $\lambda$: gradient descent step size 

\vspace{0.1cm}

    \ENSURE $\gamma = \{x_n ; n=0\ldots N\}$ minimizes $E[ \gamma ]$ conditioned on minimizing $E[ f( \gamma ) ]$

\vspace{0.1cm}

    \STATE $x_n \leftarrow \frac{N-n}{N} x_0 + \frac{n}{N} x_N \quad n \in \llbracket 0,1,\dots  N \rrbracket $ \hfill \textit{initialize with pixel-based interpolation}
   
\vspace{0.1cm}

   \STATE minimize $E[ f( \gamma ) ]$ \hfill \textit{project onto set of representational geodesics}

\vspace{0.1cm}

\WHILE{ $\gamma$ has not converged }

\vspace{0.1cm}

	\STATE $d_r \leftarrow \nabla_\gamma E[ f( \gamma ) ] $
    
\vspace{0.1cm}

\STATE $d_p \leftarrow \nabla_\gamma E[ \gamma ] $
    
\vspace{0.1cm}
    
    \STATE $\widehat{d}_{p} \leftarrow d_p - \frac{<d_r,d_p>}{ \left\Vert d_r \right\Vert^2_2 } d_r $ \hfill \textit{project out representational gradient }
    
\vspace{0.1cm}

\STATE $\gamma \leftarrow \gamma - \lambda \widehat{d}_{p} $

\vspace{0.1cm}

\STATE minimize $E[ f( \gamma ) ]$ \hfill \textit{re-project onto set of representational geodesics}
   
\vspace{0.1cm}
     \ENDWHILE

\vspace{0.1cm}

     \RETURN $\gamma$
     
\end{algorithmic}
\end{algorithm}
%It should be noted that this algorithm is generally applicable to general problems of constrained optimization. It could be applied, for example, to training a neural network while constraining it to have certain invariances. 

Despite the non-convexity of the problem, we have good reason to believe that solving this optimization problem should be feasible for trained neural networks. Since the output of the first layer is equal to the convolution of the input image with a filter bank, our problem is similar in complexity to optimizing the weights of the first layer of a network, for the same objective. Recent theoretical work shows that optimizing all layers of a network jointly makes the problem significantly more difficult than optimizing a single layer in isolation \citep{Saxe:2013tq}. Hence optimizing $E[ f( \gamma ) ]$ should be easier than training the full network for recognition. In practice we were able to solve the optimization problem for a variety of deep networks.

\begin{figure}[t]
  \begin{minipage}[c]{0.4\textwidth}
    \includegraphics[width=\textwidth]{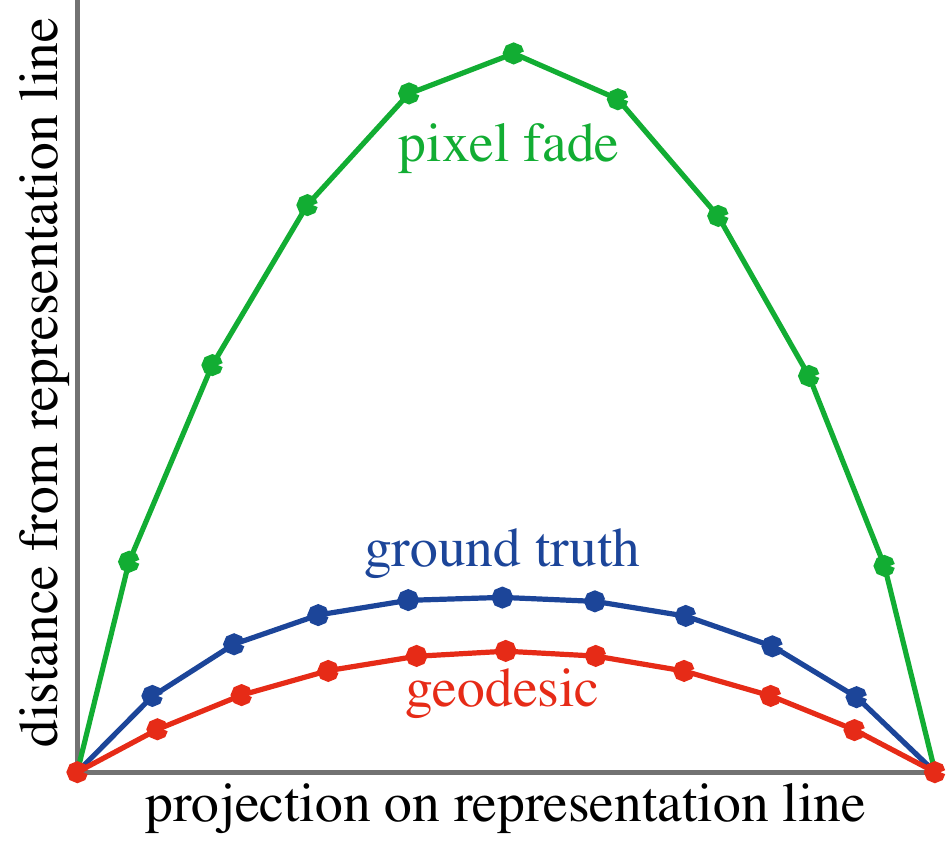}
  \end{minipage}\hfill
  \begin{minipage}[c]{0.56\textwidth}
    \caption{Deviation from the straight line connecting the representations of a pair of images, for different paths in representation space. Due to the non-linearity of the representation (the third stage of $L_2$ pooling of a deep neural network, see section \ref{sec:results}) the geodesic deviates slightly from the straight line. The ground truth transformation (here, a translation) deviates similarly, indicating that the representation has linearized the transformation to a large extent. For reference, a pixel-based interpolation deviates significantly more from a straight line. Axes are in the same units, normalized by the distance separating the end point representations. Knots along each curve indicate samples used to compute the path. 
%    The geodesic deviates slightly from the straight line, less than the ground truth transformation (here, a translation), and much less than a pixel-based interpolation. Axes are in the same units, normalized by the distance separating the end point representations. The representation used to compute the geodesic and evaluate the deviations of these paths is the third stage of $L_2$ pooling of a deep neural network (see section \ref{sec:results}).
}
\label{fig:deviation}
\end{minipage}
\end{figure}

It should be noted that if the mapping $f(\cdot)$ is not surjective, not all vectors in the representation space  are attainable from an input image. Specifically, if the mapping is non-linear (as for most representations of interest) the set of attainable vectors is non-convex, and vectors lying along the straight line connecting two representations are not necessarily attainable. As such, we can only expect to find a geodesic path whose representation is \textit{as close as possible} to this straight line by minimizing the representational geodesic cost $E[ f( \gamma ) ]$. Figure \ref{fig:deviation} shows an example of this, for the case of image translation. By construction, the geodesic is closer to a straight line in representation space than either the ground truth transformation or a pixel interpolation. The ground truth transformation lies close to the geodesic, indicating that this representation has almost (but not completely) linearized this transformation. The differences between these two paths can be made explicit by visualizing the geodesic sequence, as detailed in the following section.

\begin{figure}%[h!]
  \centering
\includegraphics[width=1\textwidth]{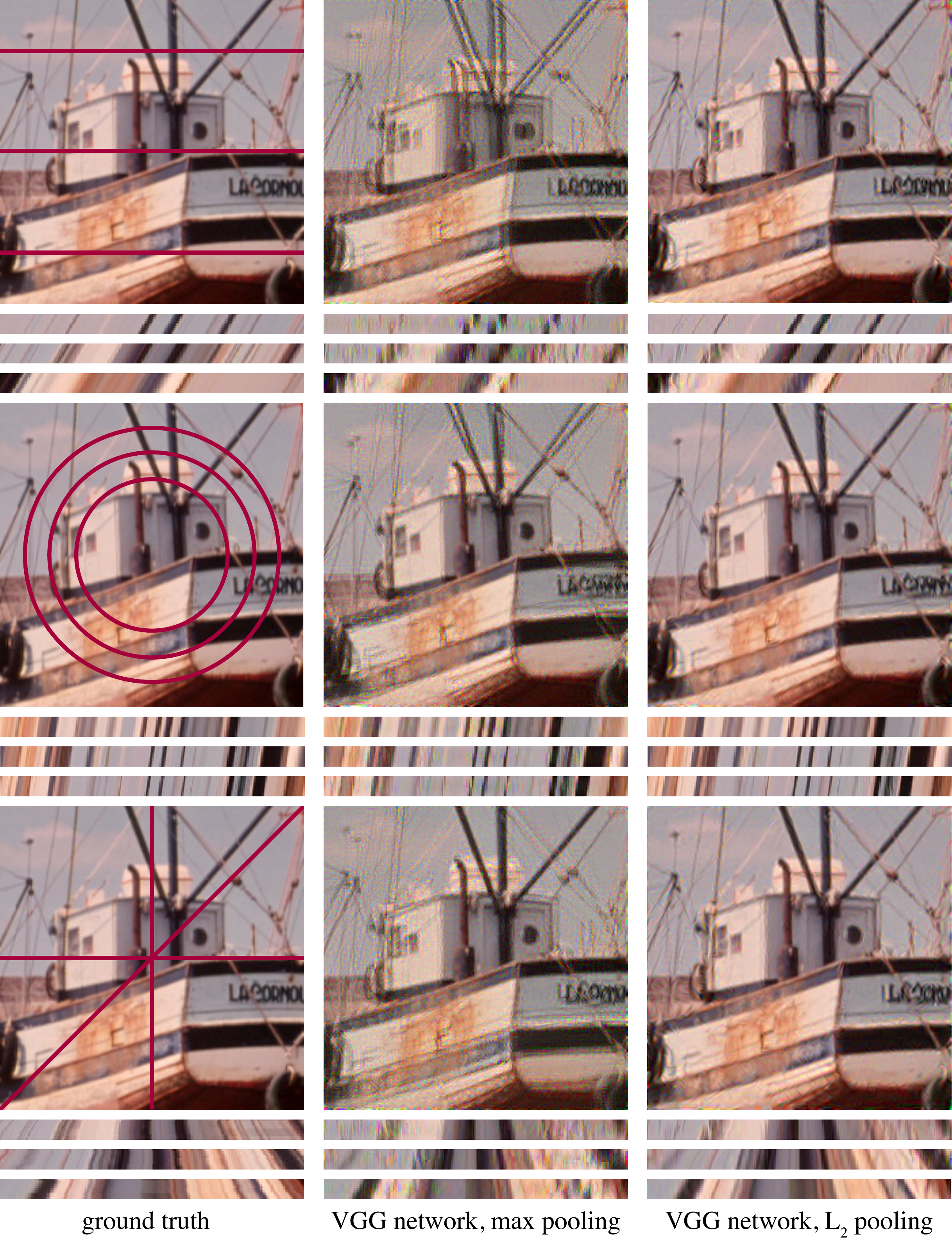}
  \caption{Comparison of geodesic sequences for VGG network representation with max pooling (middle column) and VGG network with $L_2$ pooling (right column) with ground truth sequence (left column).  Three different types of geometric transformation are tested: horizontal translation (top), rotation around the center (middle), dilation about the center (bottom).  As in figure \ref{fig:toy_example}, square images are the middle frame from the corresponding sequence, and underneath is the temporal evolution of three image slices, taken along the red lines shown in the left column. The original VGG network is unable to linearize these transformations (as indicated by the `double exposure' in the middle frame, and the discontinuous temporal slices), whereas the same VGG network with $L_2$ pooling (right column) induces a geodesic that is close to ground truth. }
  \label{fig:max_vs_l2}
\end{figure}

%%%%%%%%%%%%%%%%%%%%%%%%%%%%%%%%%%%%%%%%%%%%%%%%%%%%%%%%%%%%%%%%%%%
\section{Visualizing geodesic sequences}
\label{sec:results}
We used our geodesic framework to examine the invariance properties of the 16-layer VGG network \citep{Simonyan:2014ws}, which we chose for its conceptual simplicity and strong performance on object recognition benchmarks. As a ``representation" for our tests, we used the output of the third stage of pooling. Each stage of this continuous non-linear mapping is constructed as a composition of three elementary operations: linear filtering, half-wave rectification, and max pooling (which summarizes a local region with its maximum). We followed the preprocessing steps described in the original work: images are rescaled to the $[0, 255]$ range, color channels are permuted from RGB to BGR, and the mean BGR pixel value, $[104, 117, 124]$, is subtracted. We verified that our implementation could replicate the published object recognition results.  

\subsection{Geodesics as a diagnostic tool}

We first examined whether this representation linearizes basic geometric transformations: translation, rotation and dilation. To do so, we compute the geodesic sequence between two images that differ by one of these transformations, and compare it to the ground truth sequence obtained by incremental application of the same transformation. The extent of the overall transformation determines the difficulty of this task: all representations (even trivial ones) will produce geodesics that are close to the ground truth for very small transformations, whereas all are likely to fail for very large transformations. For our discriminative test we chose intermediate values: an 8 pixel translation, a 4$^\circ$ rotation, and a $10\%$ dilation. 

We found that the VGG network, despite its impressive classification performance, failed to linearize these simple geometric deformations and produced geodesics with salient aliasing artifacts (figure \ref{fig:max_vs_l2}, middle column). Given that no subsampling is used in the convolutional layers, we attributed this failure to the max pooling layers, which subsample the representation by a factor of 2 in each direction, despite their small spatial extent (a 2$\times$2 pooling region). To avoid aliasing artifacts when subsampling by a factor of 2, the Nyquist theorem requires blurring with a filter whose cutoff frequency is below $\frac{\pi}{2}$. Following this indication, we replaced the max pooling layers with $L_2$ pooling:
\[ 
L_2(x) = \sqrt{ g * x^2 } 
\]
where the squaring and square-root operations are point-wise, and the blurring kernel $g(\cdot)$ is chosen as a 6$\times$6 pixel Hanning window that approximately enforces the Nyquist criterion. This type of pooling is often used to describe the behavior of neurons in primary visual cortex \citep{Vintch:2015iv}, and also bears resemblance to the complex modulus used in the ``scattering transform'' \citep{Mallat:2011wz} which has been shown to be robust to smooth deformations. 

We found that this modified VGG network not only produced geodesic sequences that were free of most aliasing artifacts, but also linearized these geometric transformations convincingly, as can be seen in the temporal slices of the geodesic (figure \ref{fig:max_vs_l2}, right column). This confirms that, as with the Fourier magnitude and the scattering transform, smooth, quadratic pooling operators are able to linearize local deformations. Unlike the Fourier magnitude however, the locality and hierarchical nature of these representations tailors their invariances to a much more limited set of transformations. Furthermore, this demonstrates the power of geodesics as a visualization tool for understanding learned representations. Not only does this diagnostic report a deficiency of a representation (figure \ref{fig:max_vs_l2}, middle column), it also points to the mechanism of this failure, suggesting a simple way to improve the model. 

This suggests that the VGG network's performance on object recognition tasks could be improved by substituting max pooling with $L_2$ pooling, and retraining the network to decode this new representation. Indeed, the added invariance of this representation could enable the network to generalize to new viewing conditions more robustly.

\begin{figure}[b!]
  \centering
\includegraphics[width=1\textwidth]{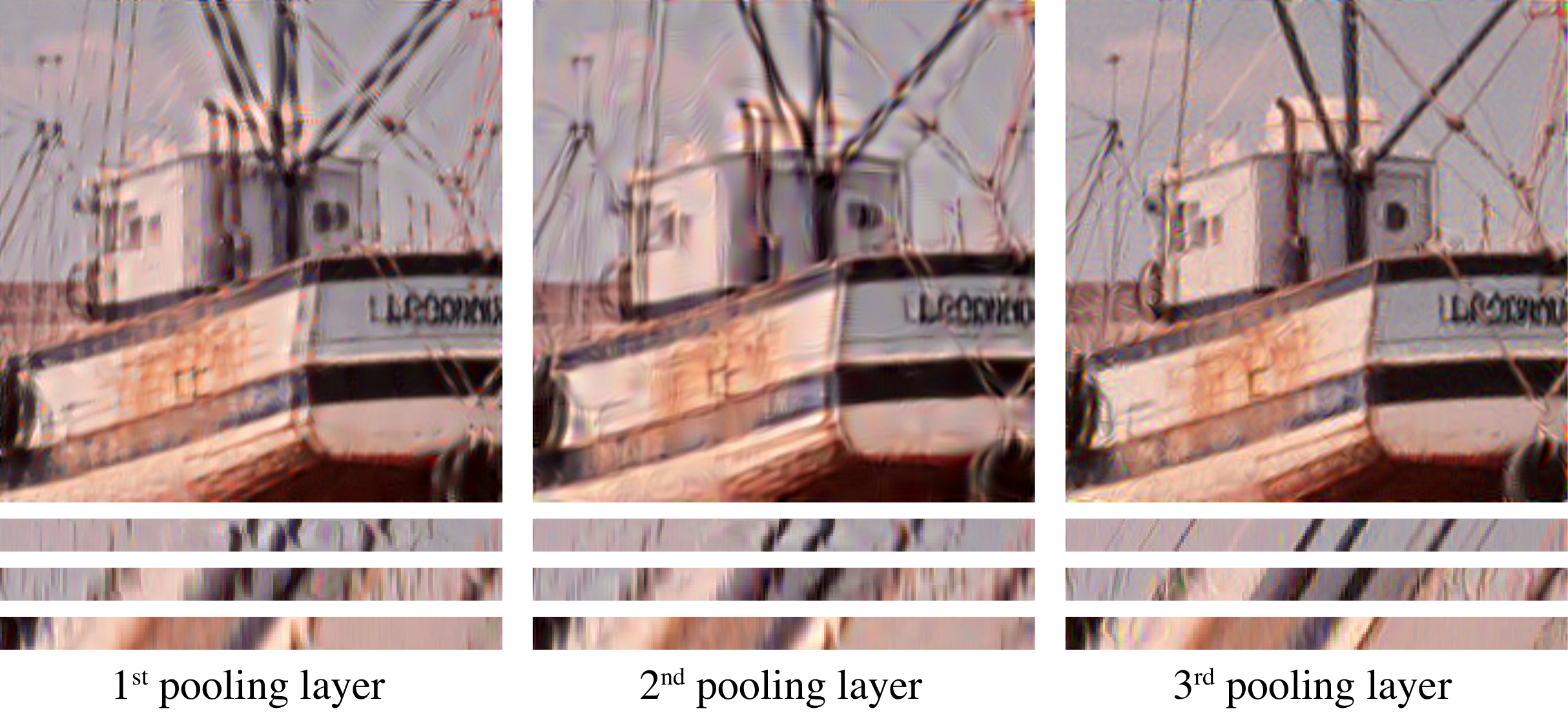}
  \caption{Even when matched for ``receptive field" size, shallow representations cannot linearize translations as well as deep ones. From left to right: geodesics generated from 1st, 2nd and 3rd pooling layers, with receptive field sizes approximately matched by altering the spatial extent of the $L_2$ pooling (to 36$\times$36, 18$\times$18 and 6$\times$6 pixels, respectively). As the complexity of the representation increases, so does the quality of the corresponding geodesic.}
  \label{fig:shallow_vs_deep}
\end{figure}

\subsection{Disambiguating spatial scale and nonlinear complexity with geodesics}
Thus far we have found that a deep representation is able to linearize a range of real-world transformations (figure \ref{fig:max_vs_l2}, right column) whereas a shallow one (e.g., the pixel intensities) is not (figure \ref{fig:toy_example}, middle column). It is unclear, however, whether the improved invariance of the deep representation is due to the spatial extent over which it computes its responses, or its nonlinear complexity. Indeed, as we progress up the hierarchy of a neural network, the effective input region for each unit (the ``receptive field") increases in size, simply due to cascaded convolution and subsampled pooling. At the same time, the complexity of the representation increases as a longer sequence of non-linear operations are composed.

In order to separate these two effects, we varied the complexity of the representation while keeping the size of the receptive field constant. For an artificial neuron, the receptive field quantifies the strength of the connection between a location in the image and that neuron's activity, and can be measured by computing the magnitude of the gradient of the neuron's activity with respect to the image. Hence, the receptive field of a non-linear neuron changes as a function of the input image. In order to measure the extent of a neuron's receptive field across all images, we averaged the magnitude of the gradient of its activity over a large set of white noise images. We generalized this method to measuring the receptive field of an entire population by computing the average magnitude of the gradient of an entire `cortical column', or set of hidden units at a given location.

Using this method, we measured the receptive field size of the representation used in our previous experiments (third pooling layer of the VGG network with $L_2$ pooling). We then computed geodesics from shallower representations (first and second pooling layers of the VGG network) for which we increased the pooling extent (from 6$\times$6 to 36$\times$36 and 18$\times$18 respectively) in order to match the receptive field size of the deep representation. These experiments show that shallower layers, despite being matched for receptive field size, are unable to linearize translations as well as deeper ones (figure \ref{fig:shallow_vs_deep}). Interestingly, we find a gradual increase in the quality of the geodesics as the complexity of the representation increases. Hence the curvature of representational geodesics, more than their dimensionality, is essential for capturing these non-linear deformations of the image. 

\begin{figure}[b!]
  \centering
\includegraphics[width=1\textwidth]{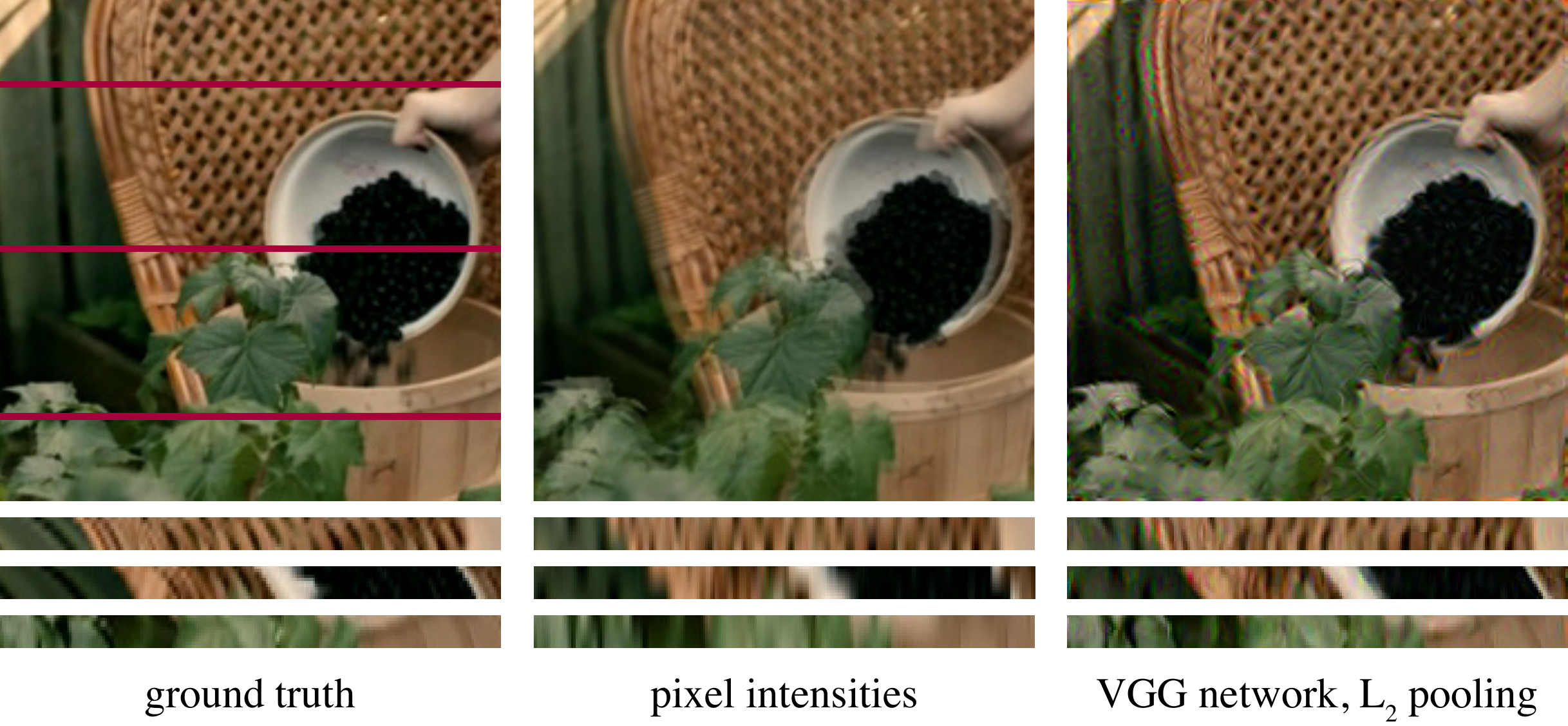}
  \caption{ Comparison of geodesic sequences for a pixel-based representation (middle column) and the VGG network with $L_2$ pooling (right column) for a natural movie (left column).  Geodesic sequences are generated between the first and last frame of the original movie.  The pixel intensity representation fails to linearize the sequence, while the VGG network with $L_2$ pooling induces a geodesic that is close to the original movie. The main deficiency in this geodesic is due to temporal aliasing, where periodic structure in the image is shifted backwards relatively to the rest of the image (see first and second temporal slices). }
  \label{fig:melancholia}
\end{figure}

\subsection{Linearizing natural image sequences}

Having tested for the modified VGG network's ability to linearize simple parametric transformations, we asked whether it can linearize compositions of these transformations that arise in natural image sequences. To explore this, we extracted 5 frames from the movie \textit{Melancholia} and generated a geodesic from the first to the last of these. We find that this geodesic smoothly transitions between the two images, and captures much of the true temporal evolution of the video (figure \ref{fig:melancholia}, left and right panels). Relative to the original sequence, the only errors it produces are due to well known problems in motion estimation. A large component of the transformation in the video is an out-of-plane rotation due to the camera panning, creating a composition of translations and dilations throughout the image. In a region of the image with periodic structure (e.g.\ the woven cane texture of the chair), the motion between the two end frames is ambiguous, because the translation between them exceeds one half of this period. This problem, known as \textit{temporal aliasing}, can be seen in the temporal slices, which reveal that the back of the chair is smoothly shifted in the opposite direction of the rest of the image (figure \ref{fig:melancholia}, right panel). In motion estimation, this problem is usually solved using a coarse-to-fine approach, in which the motion of the low frequencies is estimated first, and used to condition (or initialize) motion estimates derived from higher frequencies. This method can be naturally embedded in our framework by generalizing the nested conditionalization of geodesic objective functions (section 2.1). That is, each layer of the network can impose its own geodesic constraints, conditioned on those imposed by deeper layers. This hierarchical construction provides a means of solving the problem of temporal aliasing, and more generally should allow the network to linearize a broader class of transformations.

%%%%%%%%%%%%%%%%%%%%%%%%%%%%%%%%%%%%%%%%%%%%%%%%%%%%%%%%%%%%%%%%%%%
\section{Discussion}

\comment{***TO ADD:
%* Geodesics provide a means of visualizing metric properties of a representation.  We've devloped a methodology for generating them, and shown that...
%* can be viewed as a test of the "untangling hypothesis" proposed by di Carlo lab for biological vision: hierarchical representation aims to flatten the repreesntation of object identity, so that it can be read out with a linear decoder.
* Extensions: 
%- pairs of images with unkonwn transformation (e.g., frames from a movie)
- graded measure of invariance, as a function of extent of transformation
%- hiearchical conditional geodesics.  Adding hierarchy can serve to induce new invariances, or extend the reach of existing ones.
- connection to perception (cosyne abstract)
%- learning of representations: supervised from ground truth, or unsupervised from temporal evolution (a generalization of Slow Feature Analysis)
}

The synthesis of geodesic sequences provide a means of visualizing and assessing metric properties of a representation. We have developed a methodology for generating such sequences, and shown that they can be used as a powerful diagnostic tool for evaluating the invariance properties of learned representations. Specifically, evaluating geodesics enables one to test the ``untangling hypothesis'' by which hierarchical representations (in particular, biological sensory systems) linearize the action of identity-preserving transformations \citep{DiCarlo:2007hs}. Such an ``untangled'' representation can be linearly decoded, projecting out unwanted variations arising from image-domain renderings to achieve invariant object recognition, or projecting out the orthogonal space, so as to estimate latent rendering variables such as position, lighting, and pose. 

We used this methodology to test a state-of-the-art recognition network and found that it was unable to linearize basic image transformations such as translation, rotation and dilation. Importantly, these results suggested a simple improvement in the architecture of the network, which in turn enabled it to linearize parametric distortions as well as those found in a natural image sequence. Hence, our geodesic visualization method provides both a means of testing the untangling hypothesis for artificial networks, as well as a design tool for guiding improvements in learned representations. 

Alternatively, one could directly test the invariance of a system to a given transformation by examining the variability of responses to objects deformed by the corresponding operation. But such a test relies on establishing a meaningful measure of variability in the representation space, which is undermined by the fact that essentially equivalent representations (e.g., that differ by an invertible affine transformation) can have dramatically different distance or variability measures. As a result, it can be  difficult to compare invariance properties of different models, or even across different stages of the same network, with this direct method. The use of geodesic sequences (which are unaffected by invertible affine transformations) avoids this problem by expressing the invariance properties of the representation back in the input (image) domain, where they can be directly compared.

Moreover, our method can be applied to arbitrary image pairs, including but not limited to parametrically transformed images and frames from natural videos. For example, generating geodesics between two arbitrary images from the same object category can reveal whether object identity is an invariant of a representation. An affirmative answer implies that, back in the representation space, all of the images along the geodesic could be correctly identified using a linear decoder (as is commonly done when reading out the penultimate layer of a deep neural network).
\comment{***Modified:
, but this method is plagued with several deficiencies. First, it requires an arbitrary threshold to determine how much variability in the responses to these deformations is tolerable. Second, since the units of these representations are arbitrary, it can be difficult to compare the variability in the responses across models. Finally, it is limited to simple transformations that can be parametrically applied to a reference image. }

Finally, our method suggests a natural extension to hierarchical representations.  Our geodesic sequences were computed by minimizing path length in the pixel domain, conditioned on minimizing path length in a network representation.  This process could be applied recursively in a hierarchical representation, minimizing path length at each stage conditioned on minimal path length at higher stages. The resulting non-linear coarse-to-fine computation has the potential to solve well-known problems of temporal aliasing, and to enable hierarchical representations to linearize a much broader class of naturalistic transformations. The resulting image sequences, in turn, could be used to probe and characterize perceptual and physiological aspects of the representation of these transformations in biological visual systems.

%Finally, this method can be extended in a number of ways. First of all, generating geodesics that are close to ground truth transformations could be incorporated into an unsupervised learning framework. Indeed, given a ground truth transformation, a network could be optimized to produce geodesics that are close to this transformation. Similarly, a sequence of frames from a video could be used as the ground truth transformation connecting pairs of images, providing a simple framework for unsupervised learning from natural videos. 

%We introduce a method for synthesizing sequences of images lying along geodesics on manifolds captured by a given representation. Estimation of the perceptual distance traversed by these image sequences enables one to compare different models against each other, without worrying about the metrics in their respective domains. The model that produces the sequence with shortest perceptual length is the better model.  This method can be applied to arbitrary images pairs, including but not limited to parametrically transformed images and successive movie frames. Finally, computing geodesics on invariance manifolds exposes both insufficient and excessive amounts of invariance, as well as the curvature of the invariance manifold. Taken together, these make this method a powerful tool for measuring the invariances and selectivities of arbitrary, learned representations.

\subsubsection*{Acknowledgments}

This work was supported by the Howard Hughes Medical Institute.

%\pagebreak

\bibliography{iclr2016_conference}
\bibliographystyle{iclr2016_conference}

\end{document}